# On the Performance of an Explainable Language Model on PubMedQA

Venkat Srinivasan, Ph.D.
Vishaal Jatav, CTO
Anushka Chandrababu, Ph.D.
Geetika Sharma, MD

Gyan, Inc

**Revised: Nov 2024**

**Working Paper. Not for circulation**

**Abstract**

Large language models (LLMs) have shown significant abilities in retrieving medical knowledge, reasoning over it and answering medical questions comparably to physicians. However, these models are not interpretable, hallucinate, are difficult to maintain and require enormous compute resources for training and inference. In this paper, we report results from Gyan, an explainable language model based on an alternative architecture, on the PubmedQA data set. The Gyan LLM is a compositional language model and the model is decoupled from knowledge. Gyan is trustable, transparent, does not hallucinate and does not require significant training or compute resources. Gyan is easily transferable across domains. Gyan-4.3 achieves SOTA results on PubmedQA with 87.1% accuracy compared to 82% by MedPrompt based on GPT-4 and 81.8% by Med-PaLM 2 (Google and Deepmind). We will be reporting results for other medical data sets – MedQA, MedMCQA, MMLU – Medicine in the future.

**Introduction**

Explorations of computational methods for medical problem solving have spanned different representations and reasoning methods, including core probabilistic and decision-theoretic methods, rule-based production systems, semantic graphs, supervised learning from databases of medical information, and deep neural network models (Nori et al, 2023). These efforts have expanded from modeling image data for medical diagnostics to more general clinical reasoning using natural language processing. Recent attempts have mostly relied on deep neural networks either trained on specific medical corpora or models trained on massive amounts of general language and/or visual information and then adapted to medical data through fine-tuning.

To enable the development of more accurate models, a number of data sets have been introduced for benchmarking - PubMedQA, MedQA (USMLE), MedMCQA and MMLU among others. Models developed using domain specific data (BioLinkBert DRAGON, PubMedGPT], PubMedBERT and BioGPT (Singhal et al, 2023) have demonstrated a steady improvement in state-of-the-art (SOTA) performance on these benchmark datasets. However, larger general purpose LLMs such as GPT-3, Flan-PaLM trained on internet scale corpora with massive compute, reported significantly improved results on these data sets.

There have also been attempts to balance generalist models with domain specific data. Starting with general-purpose LLMs and then continuing to train on domain-specific data acquires a combination of both natural and domain-specific language understanding and generation skills (Gururangan et al., 2020). In the medical domain, this approach has been reported for models below 13B parameters (Lee et al., 2020; Gu et al., 2021; Peng et al., 2023; Wu et al., 2023a). At larger scales (i.e., ≥ 70B-parameters), prior studies have only explored the scope of instruction-tuning (M42-Health) or parameter-efficient finetuning (Toma et al., 2023).

Singhal at al (2023) introduced Med-PaLM 2 using a combination of an improved base LLM (PaLM 2), medical domain-specific finetuning and a novel prompting strategy that enabled improved medical reasoning. The model approached or exceeded state-of-the-art performance on MedMCQA, PubMedQA, and MMLU clinical topics datasets.

Similarly, Nori et al (2023) combined innovative prompting for GPT-4 for medical challenge problems. Their model, referred to as Medprompt, easily topped existing benchmarks for all standard medical question-answering datasets. Medprompt has outperformed state-of-the-art specialist models such as Med-PaLM 2 by large margins. On the PubMedQA data set, Medprompt reached 82% accuracy compared to 81.8% by Med-PaLM 2.

In this paper, we report results based on the Gyan-4.3 LLM which is based on a completely different architecture. Gyan-4.3 reaches SOTA results on the PubMedQA dataset.

The Gyan LLM

The Gyan LLM (Gyan) is an explainable language model constructed without reliance on large amounts of training data or probabilistic word associations. It is architecturally quite different from the neural architecture used by most large language models. Gyan is a model of language semantics. Gyan combines knowledge-based linguistics including syntactic representation, thematic roles and rhetorical structure to create a rich encoding of meaning from natural language document(s). For a detailed description of Gyan, see Srinivasan et al, 2023.

The Gyan LLM decomposes the natural language text document into an actionable meaning representation graph (GMR) which preserves the composition fully. The Gyan meaning representation contains all the concepts and relationships contained in the document at a word, clause, sentence, paragraph, discourse and document levels. Relationships between sentences, are classified by Gyan to a set of abstract rhetorical relations. The model is not a statistical model of word association but rather a compositional language model using deep semantic structural relationships.

Architecturally, therefore, the Gyan model is decoupled from data or knowledge as illustrated in Fig. 1a. With its deep linguistic pipeline, the Gyan can process any document in English (currently) out of the box. There is no training per se needed for the Gyan LLM. In contrast, pre-trained transformer-based LLMs (Neural LLMs) are next word prediction models trained on an incredibly large amount of natural language data. They are not models of language composition albeit they attempt to incorporate context in other ways.

[Revise]
Fig. 1
Gyan: High Level Architecture

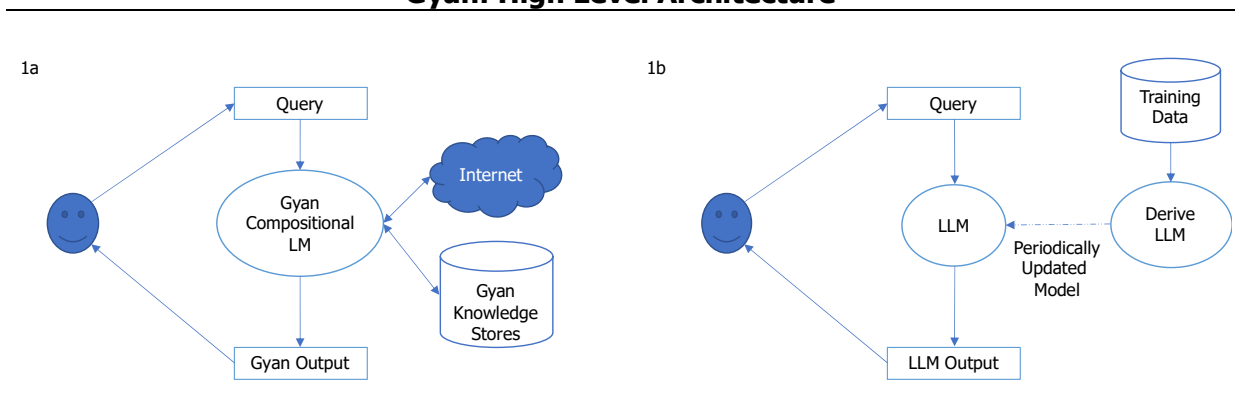

The Gyan LLM does not come built in with any domain specific knowledge. Since it is capable of encoding and decoding any document using its deep semantic model, it is transferable across domains. Gyan can be enhanced by ingesting as reference data domain specific vocabulary and knowledge, e.g., dictionaries, taxonomies like MESH, to improve its interpretation in a specific domain. From the ingested domain specific data,

Gyan creates a transparent, explainable knowledge network, 'Gyan Knowledge Net' (KN). Such a base level of knowledge is enhanced with other domain specific knowledge, e.g., PubMed.

We can view Gyan's knowledge repository in four layers as shown in Fig. 2. The first layer is factual knowledge from dictionaries, taxonomies and other factual repositories. The second level of knowledge is from expert studies most commonly in the form of academic research, e.g., PubMed. The third layer of knowledge is expert opinions, e.g., analyst reports. The final layer of knowledge can be speculative, e.g., blog posts and other social media content. All four layers together form the knowledge network for that domain.

**Fig. 2**
**Knowledge Layers in Gyan**

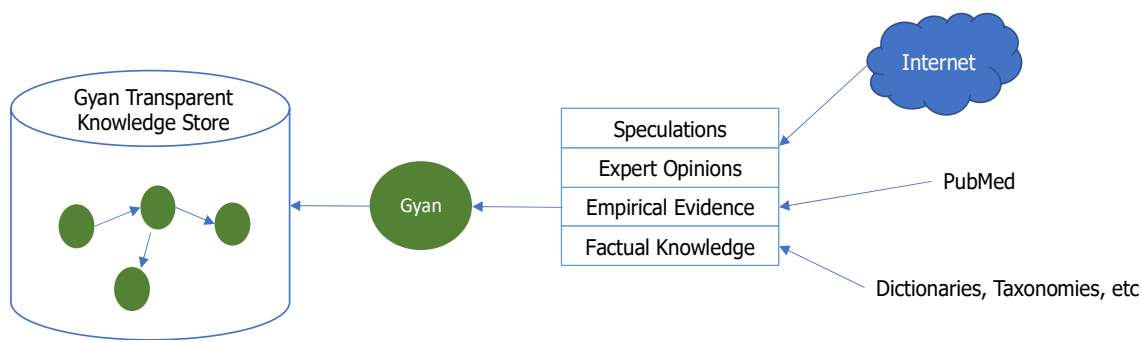

Gyan pre-processes domain knowledge content, referred to as Knowledge Stores, for fast response. Gyan can behave as a small or large language model as a function of the number of KS that are used. Gyan will offer KS based on public corpora in addition to its LLM, e.g., PubMed, Corporate SEC filings.

Gyan can also acquire knowledge dynamically from the Internet. In the case of the internet, Gyan LLM retrieves results using search engines and processes them real time. Thus, asymptotically, the amount of knowledge accessible to Gyan and neural LLMs will be the same if **a set of non-overlapping sources** of training data used to train the neural LLMs are used to create Gyan 'knowledge stores'. It is important to note that Gyan will only need a non-overlapping set of knowledge sources. The Gyan LLM is not using frequency of patterns or based on probabilistic word associations.

Because Gyan's meaning representation graphs are a faithful representation of the underlying document and completely invertible, Gyan is fully explainable and traceable, will not hallucinate, cannot be manipulated or misused with data and is transferable across domains easily. Gyan can guarantee that enterprise data will never be co-mingled with external data and vice-versa. Since the Gyan LM does not require any training, it needs a fraction of the compute resources neural LLMs need for training and inference.

With the pre-training architecture, neural LLMs have to be periodically updated with new data. Another serious limitation of the pre-trained transformer architecture is that this can cause serious instability or catastrophic forgetting [Kwiatkowski, 2019, Li et al, 2022, Zhai et al, 2024]. It is possible that for the same query, the neural LLM might produce an entirely different or incorrect or misleading response after an update. While some attempts have been made to limit such 'catastrophic forgetting' [Wang, et al, 2024], the issue is far from being solved.

In contrast, Gyan is easily maintained. With its decoupled knowledge architecture, Gyan is able to reflect new knowledge continuously and does not suffer from the 'catastrophic forgetting' phenomenon. New knowledge can be maintained independent of existing knowledge or merged as needed.

An important step in the Gyan reasoning framework is the expansion of local context to a more global context using its deep semantic model and its KS. Human understanding of natural language is a combination of two essential elements – the query or document on the one hand and pre-existing human knowledge on the other. Humans interpret or understand a document in the context of what we know already or our pre-existing cognitive perception (DiMaggio, 1997; Paller, 2002).

In the case of a query, humans expand it to create an expanded context in their minds. A similar process occurs with a document we read. We expand it in our mind with all the knowledge we already know about various concepts and relationships expressed in the document. Gyan attempts to mimic the human process by automatically expanding the queries and documents. Gyan constructs a global context ("world model") for every query and document leveraging its KN and applicable KS.

Gyan looks for related concepts outside of the current document which might have a bearing on the meaning representation for the current document. Depending on the user's preference, Gyan can use just its factual knowledge store, KN, and/or expand it to include additional public or private KS. The expanded GMR allows Gyan to establish semantic relationships between concepts in the document which cannot be inferred from processing the document alone. Humans reading the document would have readily made the same association with their prior knowledge.

**Gyan and PubMedQA**

**The Dataset**
While many large-scale annotated general domain QA datasets have been introduced (Rajpurkar et al., 2016; Lai et al., 2017; Kocisk et al., 2018; Yang et al., 2018; Pampari et al., 2018; Pappas et al., 2018; Kim et al., 2018; Kwiatkowski et al., 2019), their questions are mostly simple factual questions whose answers can be extracted in the contexts without much reasoning.

Jin et al (2019) created the PubMedQA dataset to provide a sizable biomedical dataset which (1) has substantial instances with some expert annotations and (2) requires reasoning over the contexts to answer the questions. They found that around 760k articles in PubMed use questions as their titles. Among them, the abstracts of about 120k articles are written in a structured style – with subsections of "Introduction", "Results" etc. Conclusive parts of the abstracts, often in "Conclusions", are the authors' answers to the question title. Other abstract parts can be viewed as the contexts for giving such answers. Interestingly, more than half of the question titles of PubMed articles can be briefly answered by yes/no/maybe, which is significantly higher than the proportions of such questions in other datasets, e.g.: just 1% in Natural Questions (Kwiatkowski et al., 2019) and 6% in HotpotQA (Yang et al., 2018).

Jim et al (2019) collected all PubMed articles with question titles, and manually labeled 1k of them for cross-validation and testing. The rest of yes/no/answerable QA instances compose of the unlabeled subset for semi-supervised learning. They also automatically converted statement titles of 211.3k PubMed articles to questions and label them with yes/no answers using a simple heuristic so such artificially generated instanced can be used for pre-training. In PubMedQA contexts are generated to answer the questions and both are written by the same authors. This consistency assures that contexts are perfectly related to the questions and is one of the reasons why it is a widely accepted dataset for benchmarking purposes for testing scientific reasoning.

Based on an analysis of 200 examples from the 1k labeled dataset, Jin et al (2019) summarize the type of questions and types of reasoning required to answer them, which is reproduced in Fig.3.

The main metrics of PubMedQA are accuracy and macro-F1 on PQA-L test set using question and context as input. We denote prediction using question and context as a reasoning-required setting, because under this setting answers are not directly expressed in the input and reasoning over the contexts is required to answer the question.

A parallel setting, where models can use question and long answer to predict yes/no/maybe answer, is denoted as reasoning-free setting since yes/no/maybe are usually explicitly expressed in the long answers (i.e.: conclusions of the abstracts). Obviously, it's a much easier setting which can be exploited for bootstrapping PQA-U (Jin et al, 2019).

**Fig. 3
Question and Reasoning Type in PubMedQA (Reasoning Required?)**

| Question Type | % | Example Questions |
|---|---|---|
| Does a factor *influence* the output? | 36.5 | Does reducing spasticity *translate into* functional benefit? Does ibuprofen *increase* perioperative blood loss during hip arthroplasty? |
| Is a therapy *good/necessary*? | 26.0 | *Should* circumcision be performed in childhood? Is external palliative radiotherapy for gallbladder carcinoma *effective*? |
| Is a *statement* true? | 18.0 | Sternal fracture in growing children: *A rare and often overlooked fracture?* Xanthogranulomatous cholecystitis: *a premalignant condition*? |
| Is a factor *related to* the output? | 18.0 | Can PRISM *predict* length of PICU stay? Is trabecular bone *related to* primary stability of miniscrews? |
| **Reasoning Type** | **%** | **Example Snippet in Context** |
| *Inter-group* comparison | 57.5 | [...] Postoperative AF was significantly lower in the *Statin group* compared with the *Non-statin group* (16% versus 33%, p=0.005). [...] |
| Interpreting *subgroup* statistics | 16.5 | [...] 57% of patients were *of lower socioeconomic status* and they had more health problems, less functioning, and more symptoms [...] |
| Interpreting *(single) group* statistics | 16.0 | [...] *A total of 4 children* aged 5-14 years with a sternal fracture were treated in 2 years, 2 children were hospitalized for pain management and [...] |
| **Text Interpretations of Numbers** | **%** | **Example Snippet in Context** |
| Existing *interpretations* of *numbers* | 75.5 | [...] Postoperative AF was *significantly lower* in the Statin group compared with the Non-statin group (*16% versus 33%, p=0.005*). [...] |
| No interpretations (*numbers only*) | 21.0 | [...] 30-day mortality was *12.4%* in those aged<70 years and *22%* in those>70 years (*p<0.001*). [...] |
| No numbers (*texts only*) | 3.5 | [...] The halofantrine therapeutic dose group showed *loss and distortion of inner hair cells and inner phalangeal cells* [...] |

Table 3: Summary of PubMedQA question types, reasoning types and whether there are text descriptions of the statistics in context. Colored texts are matched key phrases (sentences) between types and examples.

All the above question and reasoning types are part of the abstract relationships included in Gyan-4.3.

**Results and Observations**

In order to process the PubMedQA dataset, we created an initial Gyan Knowledge Network (KN) for medicine with medical dictionaries, MESH and medical text books. We have so far ingested 32 different domain specific datasets as reference data. See Appendix B for a description of these data sets. In terms of base level metrics, this yielded in 200k concepts and approximately 89 million relations. As mentioned previously, Gyan constructs its KN from this reference data.

The PQA-L dataset of 1K questions and abstracts has been further divided into 500 for training and 500 for validation. For a detailed description of the PubMedQA data set, pl see Jin et al (2019).

Based on the initial KN for Medicine, we processed the Test Set comprising 500 questions with Gyan-4.3. The results are shown in Fig. 5.

**Fig. 5**
**Gyan-4.3 on the Test Set**

Test Set

| Category | Count | Percentage |
|---|---|---|
| Correct | 418 | 87.08% |
| Incorrect | 62 | 12.92% |
| Total | 480 | 100.00% |

Out of the 500, 20 instances in the dataset were incomplete.

Gyan-4.3 answered 87.08% of the questions correctly. As we can see from the table and from the leaderboard [https://pubmedqa.github.io/], with the initial KN, as of Nov 30, 2024, Gyan-4.3 has already reached SOTA performance levels.

The Gyan LLM's reasoning framework is illustrated in Fig. 7 for a PubMedQA question from the test set. Each question is associated with a PubMed abstract and the LLM has to demonstrate multi-layer reasoning to answer the question. Each abstract is divided into a Purpose, Methods and Results and Conclusions section. The LLM is required to ignore the Conclusions section and determine the response to the question from Results.

**Fig. 7**
**Gyan Reasoning: An Illustrative Example**

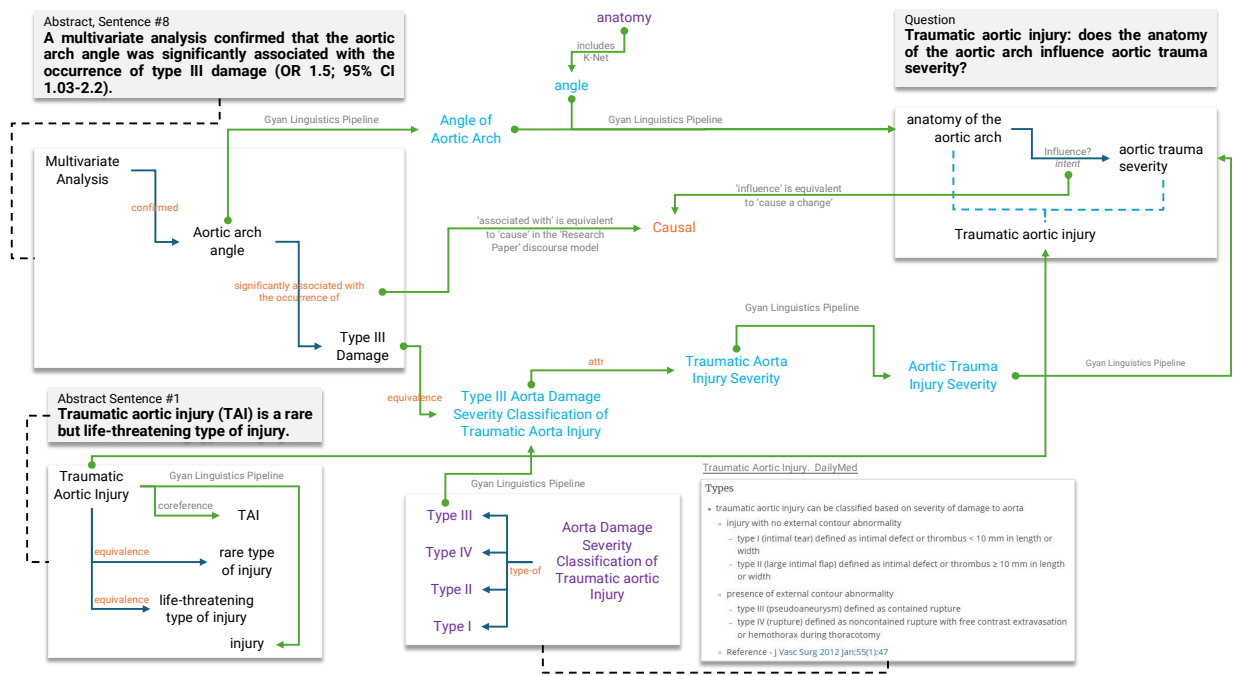

Gyan performed a series of inferences some of which were part of its deep semantic language model and others were the use of knowledge from the ingested KS for Medicine. Fig. 7 illustrates the Gyan reasoning and inferencing process from the Results section to the Query section.

Gyan determined that sentence #8 [S8] contains the answer to the Question through its reasoning and inference.

1. Gyan determines Aortic Arch Angle in S8 is the same as 'Anatomy of the Aortic Arch' in the question. This derivation is explained in Fig. 7. The determination involves both aspects of the Gyan language model and expansion of knowledge using Wikipedia and other sources identified in Fig. 7.

2. Similarly, Gyan determines that Type III damage is equivalent in meaning to Aortic Trauma Severity. Gyan relies on a definition of Severity Classification of Traumatic Aortic Injury provided in DailyMed to determine that Type III damage was referring to aortic trauma severity.

3. Gyan further determines that 'significantly associated with the occurrence of' to be a causal expression. The rhetorical expression 'influence' in the question is also inferred to belong to the Gyan abstract relation type 'causal'.

With Steps 1-3, Gyan can assert that S8 answers the question affirmatively.

As mentioned earlier, Gyan is fully tractable. Improving performance requires expanding the initial KN with missing knowledge. While it is possible that may be there are some unique discourse model elements manifested in the question and reasoning types in Fig. 3, it is unlikely. As we improve the KN for Medicine, there will be no catastrophic forgetting. Even if there is a conflict in some knowledge as applied to a specific context and because of this there is a contradictory answer to one or more questions, it will be completely transparent and the ambiguity in the corresponding knowledge representation can be easily resolved with additional knowledge.

We also applied Gyan-4.3 to the complete 1k dataset. Recall that Gyan-4.3 was not trained using the training set. The results are shown in Fig. 8.

**Fig. 8**
**Full Labeled Dataset**

| Category | Count | Percentage |
|---|---|---|
| Correct | 819 | 85.58% |
| Incorrect | 138 | 14.42% |
| **Total** | **957** | **100.00%** |

Out of the 1,000, 43 instances in the dataset were incomplete

We can see that the results overall are reasonably close to the performance on the test set. This suggests that the Gyan-4.3 KS for Medicine is reasonably complete with respect to the knowledge required for the PubMedQA data set. The only reason for Gyan-4.3 not to be at 100% accuracy is the missing knowledge for some questions. We can easily identify the knowledge that is missing and add it to the KS.

**Discussion & Future Work**

In this paper, we have illustrated the efficacy of a novel alternative architecture to neural LLMs on the PubMedQA dataset. Gyan-4.3 reached SOTA performance on the PubMedQA dataset with a Knowledge Store for Medicine comprising knowledge gleaned from 32 different domain specific sources. The results demonstrate that the alternative architecture can be as effective as neural LLMs without any of the challenges of the neural LLM architecture.

We are now enhancing the KS for Medicine by pre-processing medical text books provided with MedQA and others available in the public domain. We anticipate enhanced KS for Medicine further improving Gyan performance on PubMedQA. Work is also underway to evaluate the efficacy of the Gyan architecture on MedQA, MMLU – Medicine and other non-medicine datasets. We will be reporting these results in the future.

# Appendix I
# PubMedQA Leaderboard with Gyan-4.3 [as of Nov 15, 2024?]

| Rank | Date | Model | Accuracy |
|---|---|---|---|
| **1** | **Oct 30, 2024** | **Gyan-4.3** | **87.10%** |
| | | *Gyan* | |
| **2** | Nov 28, 2023 | **GPT-4 (Medprompt)** | 82.00% |
| | | *Microsoft* | |
| **3** | May 16, 2023 | **Med-PaLM 2** | 81.80% |
| | | *Google Research & Deepmind* | |
| **4** | Nov 27, 2023 | **MEDITRON** | 81.60% |
| | | *EPFL* | |
| **5** | Jul 6, 2023 | **Palmyra-Med** | 81.10% |
| | | *Ant Group* | |
| **6** | Dec 1, 2023 | **AntGLM-Med** | 80.60% |
| | | *Ant Group* | |
| **7** | Apr 12, 2023 | **GPT-4-base** | 80.40% |
| | | *Microsoft & OpenAI* | |
| **8** | Mar 4, 2024 | **Claude-3** | 79.70% |
| | | *Anthropic* | |
| **9** | Jan 11, 2023 | **GPT 3.5 + Z-Code++** | 79.60% |
| | | *Microsoft Azure AI* | |
| **10** | Dec 26, 2022 | **Flan-PaLM(3-shot)** | 79.00% |
| | | *Google Research & Deepmind* | |
| **11** | Mar 14, 2024 | **HEAL** | 78.40% |
| | | *DeepScribe Inc* | |
| **12** | Dec 20, 2022 | **Codex (5-shot)** | 78.20% |
| | | *Technical University of Denmark & Copenhagen University Hospital* | |
| **13** | Sep 13, 2019 | **Human Performance** | 78.00% |
| | | *University of Pittsburgh & Carnegie* | |
| **14** | Nov 16, 2022 | **Galactica** | 77.60% |
| | | *Meta AI* | |
| **15** | May 22, 2023 | **GatorTronGPT** | 77.60% |
| | | *University of Florida & NVIDIA* | |
| **16** | Mar 4, 2024 | **MedSwift-XL** | 76.80% |
| | | *Cerebras Systems* | |
| **17** | Mar 20, 2023 | **GPT-4** | 75.20% |
| | | *Microsoft & OpenAI* | |
| **18** | Apr 18. 2024 | **Reka Core** | 74.60% |
| | | *Reka* | |
| **19** | Dec 15, 2022 | **PubMedGPT** | 74.40% |
| | | *Stanford University* | |
| **20** | Oct 17, 2022 | **DRAGON** | 73.40% |
| | | *Stanford University & EPFL* | |

# Appendix II
# List of Vocabulary Resources Ingested

| Source | Type of Information Provided | Data types covered |
|---|---|---|
| Cleveland Clinic | Clinical guidelines, disease symptoms, treatment options, patient education materials. | Condition Prevalence, Condition Symptoms, Condition Complications, Condition Risk Factors, Condition Treatment, Condition Laboratory Findings, Condition Imaging Finding, Condition Therapy, Condition Treatment Guidelines, Condition Prognostic Factors, Condition Subtype, Disease/condition Pathogen, Disease Target, Drug Avoid, Drug Mechanism of Action, Drug SideEffects, Drug Symptom/Disease, Drug Target, Device Contra, Device Symptom/Disease, Epidemiology Parameter description, GeneticResult Significance, TestAssessment Finding, TestAssessment Pathology, Treatment Mechanism Action, Treatment Risk Reduction, RiskFactor Details, Microorganism Symptoms, Microorganism Condition, Surgery Organ, Procedure Drug, Condition Biological Mechanism |
| nhsinform-symptoms | Symptoms of diseases, healthcare guidelines. | Condition Symptoms, Condition Complications, Condition Risk Factors, Condition Treatment, Condition Therapy, Condition Treatment Guidelines, Condition Prognostic Factors, Condition Prevalence, Disease/condition Pathogen, Drug Symptom/Disease, TestAssessment Finding, Microorganism Symptoms, Microorganism Condition |
| OrphaNet | Rare diseases, genetic disorders, orphan drug information, and clinical trial references. | Condition Subtype, Condition Complications, Condition Therapy, Condition Treatment Guidelines, GeneticResult Significance, Condition Genetic Mechanism (related to rare diseases), Condition Inheritance Pattern |
| Mayo Clinic | Evidence-based disease overviews, diagnostic procedures, treatment options, lifestyle recommendations. | Condition Symptoms, Condition Complications, Condition Risk Factors, Condition Treatment, Condition Laboratory Findings, Condition Imaging Finding, Condition Therapy, Condition Treatment Guidelines, Disease Target, Drug Symptom/Disease, TestAssessment Finding, Microorganism Symptoms, Procedure Drug |
| RxList | Drug information, side effects, dosage, drug interactions, brand vs. generic drug details. | Drug Class, Drug Component, Drug Contra, Drug Mechanism of Action, Drug Side Effects, Drug Symptom/Disease, Drug Target, DrugInteraction Symptom |
| Medline | Medical research articles, drug information, disease management guidelines, health literacy materials. | TestAssessment Finding, TestAssessment Pathology, Diagnostic test parameter Effect, Condition Test Assessment, Condition Laboratory Findings, Drug Avoid (as related to test preparations or contraindications), Procedure Drug (related to diagnostic procedures), Microorganism Condition (as identified by specific tests), TestAssessmentResult Interpretation |
| dictionary-apa-org | Psychological and mental health terminology definitions, APA-referenced concepts. | Concept Definition |
| bv-brc-org | Microrganism related information | Microorganism Features, Microorganism Category, Microorganism Condition, Microorganism Symptoms, Pathogen MechanismAction, Cellular Process Mechanism, Process Immune Mechanism, Pathway Mechanism |
| medicine-guide | General medical guidelines, including drug prescriptions and disease management. | Condition Treatment, Condition Therapy, Condition Surgical Options, Device Treatment, Disease Targeted Therapies, Procedure Specific Surgeries, Treatment Surgical Methods |
| dailymed-drug-classes | FDA-approved drug labels, medication usage, pharmacological classifications, regulatory information. | Drug Class, Drug Mechanism of Action, Drug Physiologic Effect, Drug Chemical/Ingredient |
| aclandanatomy | Human anatomy resources | Anatomy Organ/Structure, Anatomy Function, Anatomy Embryologic Origin, Condition Anatomical Structure Involved, CellType Organelle Location |
| merriam-webster-medical | Medical terminology definitions, layman-friendly medical explanations. | Concept Definition, Condition symptom, Drug Class, Drug Contra, Procedure Condition, Treatment Condition, Symptom Condition |
| orpha-net-intellectual-disability | Specific focus on intellectual disabilities, genetic causes, diagnostic guidelines, and treatment. | GeneticResult Significance, GeneticTerm Definition, Gene Pathology, Condition Genetic Mechanism, Condition Inheritance Pattern, Disease Target |
| nhsinform-procedures | Details about medical procedures, preoperative and postoperative care, patient instructions. | Procedure Drug, Surgery Organ, Condition Surgical Options, Treatment Surgical Methods |
| oxfordreference | General medical and healthcare-related reference materials, terminology, and explanations. | Epidemiology Parameter description, Health Economics Concept, Condition Prevalence, Condition Risk Factors, Condition Category, StatisticsTerm Implications, Study Population, Study Association, Study design Limitation |
| Drug target - Data Bank | Information about pharmacological targets, mechanisms of action, molecular pathways of drugs. | Drug Target, Pathway Mechanism, Drug Mechanism of Action, Condition Therapy, Treatment Mechanism Action |
| pressbooks-dev | An educational repository with medical and healthcare books or open-access course materials. | Anatomy Organ/Structure, Anatomy Function, Anatomy Embryologic Origin, Condition Anatomical Structure Involved, Condition Anatomical Target, CellType Organelle Location, Organelle Function, Muscle Condition, Epithelium Location |
| Pathology Dictionary | Detailed definitions and explanations to understand various pathology terms and phrases | Pathology Details, CellType Organelle Location, Condition Pathology, TestAssessment Pathology, Gene Pathology, Disease Pathology, Symptom Pathology |
| nhs-conditions | Comprehensive information about various health conditions, including their symptoms, treatments, and complications | Condition Symptoms, Condition Complications, Condition Risk Factors, Condition Treatment, Condition Therapy, Condition Surgical Options, Disease/Condition Pathogen, Treatment Surgical Methods, Drug Symptom/Disease |
| Drugs and Supplements | Comprehensive information on various drugs and supplements, including their uses, side effects, and interactions. | Drug Class, Drug Mechanism of Action, Drug Side Effects, Drug Contra, Drug Symptom/Disease, DrugInteraction Symptom |
| mayoclinic-tests-procedures | Comprehensive details about various medical tests and procedures, | TestAssessment Finding, TestAssessment Pathology, TestAssessmentResult Interpretation, Procedure Drug, Condition Test Assessment, Diagnostic test parameter Effect, Process Outcome, Surgery Organ |
| bv-brc-virus | Genetic and molecular information about various viruses, their mechanisms of action, associated conditions, and immune processes | Microorganism Features, Microorganism Condition, Pathogen MechanismAction, Disease/Condition Pathogen, Cellular Process Mechanism, Process Immune Mechanism |
| nci-genetics-dictionary | Provides definitions and explanations for a range of genetics-related terms, which is useful for understanding complex genetic conditions and research. | Gene Pathology, GeneticResult Significance, GeneticTerm Definition, Condition Genetic Mechanism, Disease Target, Cellular Process Mechanism |
| healthdirect-procedures | Comprehensive information on various medical procedures | Procedure Drug, Surgery Organ, Condition Surgery, Condition Therapy, Diagnostic test parameter Effect, TestAssessment Finding, Process Outcome |